\documentclass[11pt]{article}

\usepackage[preprint]{acl}

\usepackage{microtype}
\usepackage[utf8]{inputenc}
\usepackage{times}
\usepackage{geometry}
\usepackage{titlesec}
\usepackage{graphicx}
\usepackage{booktabs}
\usepackage{array}
\usepackage{float}
\usepackage{url}
\usepackage{hyperref}
\usepackage{tcolorbox}
\usepackage{xcolor}
\usepackage{subcaption}
\usepackage{booktabs}
\usepackage{placeins}
\usepackage{multirow}
\usepackage{devanagari}
\usepackage{natbib}
\usepackage[english,bidi=default]{babel} 

\titleformat{\section}{\bfseries\large}{\thesection.}{0.5em}{}
\titleformat{\subsection}{\bfseries\normalsize}{\thesubsection.}{0.5em}{}

\begin{document}

\title{\textbf{Small Wins Big: Comparing Large Language Models and Domain Fine-Tuned Models for Sarcasm Detection in Code-Mixed Hinglish Text}}

\author{
  Bitan Majumder\\
  \texttt{Pondicherry University}\\
  \texttt{Email}
  \and
  Anirban Sen\\
  \texttt{Ashoka University}\\
  \texttt{Email}
}
\author{Bitan Majumder \\
  Pondicherry University \\
  \texttt{bitanmajumder2002@gmail.com} \\\And
  Anirban Sen \\
  Ashoka University  \\
  \texttt{anirban.sen@ashoka.edu.in} \\}

\date{}

\maketitle

\begin{abstract}
Sarcasm detection in multilingual and code-mixed environments remains a challenging task for natural language processing models due to structural variations, informal expressions, and low-resource linguistic availability. This study compares four large language models, Llama 3.1, Mistral, Gemma 3, and Phi-4, with a fine-tuned DistilBERT model for sarcasm detection in code-mixed Hinglish text. The results indicate that the smaller, sequentially fine-tuned DistilBERT model achieved the highest overall accuracy of 84\%, outperforming all of the LLMs in zero and few-shot set ups, using minimal LLM generated code-mixed data used for fine-tuning. These findings indicate that domain-adaptive fine-tuning of smaller transformer based models may significantly improve sarcasm detection over general LLM inference, in low-resource and data scarce settings.
\end{abstract}


\section{Introduction}
Sarcasm is a style of figurative speech characterized by deliberately stating something that is not literally true, and it reflects the complexity of social communication beyond literal meaning \cite{mcauley2025humor}. Sarcasm plays a crucial role in online communication. Many users post messages on social media that are sarcastic and are intended to convey their position on specific socio-political issues \cite{Alqahtani2023}. The presence of sarcasm is also evident across other web-based sources, including blogs, multimodal audio and video content, and digital news media. Sarcastic messages often attract greater attention than their non-sarcastic counterparts that express similar or equivalent viewpoints. Consequently, accurate sarcasm detection is an important task that can substantially enhance discourse analysis in online media and other web-based data.

However, the presence of sarcasm has made online discourse analysis a significantly challenging task. Standard natural language processing (NLP) techniques often fail to recognize sarcasm, thereby producing erroneous results and misleading statistics in the analysis of online discourse. These techniques typically assume the surface-level literal meaning of sarcastic text, leading to an incorrect interpretation of the underlying message. With the advent of large language models (LLMs), this problem has been partially alleviated. Owing to their pre-training on large-scale world knowledge, LLMs demonstrate improved performance in sarcasm detection compared to traditional methods. Nevertheless, a substantial research gap remains, as even state-of-the-art LLMs continue to perform inadequately on sarcasm detection tasks \cite{zhang2025sarcasmbench}.

The problem is further compounded when operating on code-mixed messages. Code-mixing refers to the linguistic phenomenon in which elements from two or more languages or linguistic varieties are combined within a single conversation, sentence, or phrase. An example of a code-mixed sentence is "{\dn y\?} roads {\dn aApk\?} electric car {\dn ko} drive {\dn krt\?} {\dn smy} recharge {\dn kr} {\dn skt\?} {\dn h\?}" (meaning ``These roads can charge your electric car while driving"). A large number of social media users employ code-mixed language to express their positions, particularly in the Indian context. Sarcasm detection becomes especially challenging in code-mixed settings such as Hinglish (Hindi–English), where grammatical structures, script usage, and lexical patterns vary fluidly. Additionally, the availability of code-mixed text data on sarcasm is severely limited, leading to challenges in training/fine-tuning relevant models. The performance of most traditional methods and LLMs in code-mixed sarcasm detection thus remains dissatisfactory. Thus, we intend to answer the following research question through this study: \textit{Do domain fine-tuned classical models exhibit performance comparable to LLMs in code-mixed sarcasm detection, when fine-tuned on a limited amount of code-mixed data?}

Prior work has predominantly focused on sarcasm detection in English, with relatively limited research addressing multilingual or code-mixed settings \cite{katyayan2019sarcasm}. In this paper, we evaluate how classical transformer-based models perform on sarcasm detection in Hinglish sentences (Hindi+English code-mixed sentences), and examine whether their performance is comparable to that of LLMs in zero-shot and few-shot settings. Specifically, we employ a domain fine-tuned transformer-based classifier (DistilBERT) for sarcasm detection in Hinglish text. We then compare its performance against four state-of-the-art LLMs (Llama 3.1, Mistral, Gemma 3, and Phi 4) evaluated under zero-shot and few-shot prompting settings. In addition, we assess the performance of the classical model in a transfer learning setup, where sentiment detection is used as an auxiliary task.

To address the challenge of unavailability of equivalent code-mixed sarcasm data, we leverage a SOTA LLM (Gemini 2.5 Pro) to translate a subset of original sentences to their code-mixed counterparts. We leverage this synthetically generated data for fine-tuning.

Our findings indicate that the smaller, domain fine-tuned model (trained on English data, fine-tuned on minimal code-mixed data) substantially outperforms all four LLMs in code-mixed sarcasm detection, achieving improvements of over 20 percentage points in both accuracy and F1 score across both zero-shot and few-shot settings. This performance gain is particularly remarkable given the limited amount of code-mixed data only used for fine-tuning. Furthermore, the sentiment fine-tuned model also surpasses most LLMs in terms of both accuracy and F1 score, highlighting the effectiveness of task-specific and transfer learning approaches around smaller, classical models for sarcasm detection in code-mixed text.

The contributions of this study are twofold: (A) We release an LLM generated Hinglish sarcasm detection dataset that can support further research in this area as well as related downstream tasks (the sample of the dataset could be found in Appendix \ref{appendix:example_data}), and (B) we demonstrate the superiority of minimally domain fine-tuned classical models compared to state-of-the-art (SOTA) LLMs, given the dataset used. Our approach of using a classical domain fine-tuned model is suitable for deployment in low-infrastructure and resource-constrained settings, for the challenging task of sarcasm detection in code-mixed text.

\section{Related Work}
Sarcasm detection is a much studied area of research \cite{yacoub2024survey,ganganwar2024sarcasm}. Previous studies have focused on employing classical machine learning \cite{vsandor2024sarcasm,prajapati2024sarcasm} and transformer based approaches \cite{khan2025novel,helal2024contextual}, and exhibiting their efficiency in the task of sarcasm detection.

Sarcasm in text frequently co-occurs with additional microblog features like emojis, and often has dependency on sentence context. In this direction, \cite{kumar2023hybrid} demonstrate a CNN and LSTM based deep learning approach that captures sarcasm in text data through text and emoji embeddings. \cite{helal2024contextual} propose a transformer based approach to use the context of the sentence in detecting sarcasm. Additionally, sarcasm in publicly available web-based data (news, social media) is not merely limited to text, but multimodal content. \cite{jia2024debiasing} use contrastive learning in a transformer based set up (RoBERTa and Vision Transformer) to detect sarcasm in multimodal content, with synthetic data augmentation using an LLM. In a similar study, \cite{liu2024sarcasm} employ a sentiment-aware attention mechanism and combine image and textual features to compute a Sentiment-Aware Text-Image Contrastive Loss, enabling sarcasm detection. 

While a significant amount of work exists in the space of monolingual sarcasm detection, a few studies have focused on code-mixed content. Sarcasm detection in code-mixed data is a significantly challenging problem, primarily owing to the lack of annotated datasets, the complexity of the language mixing, and the role of contextual cues \cite{ojha2026sarcasm}. In this direction, \cite{pandey2023bert} employ an approach to detect sarcasm in code-mixed data based on a BERT embedding model and an LSTM classifier. \cite{pandey2025hybrid} propose a hybrid CNN based model that integrates character and word embeddings to detect multilingual sarcastic content. \cite{ratnavel2023sarcasm} propose a transformer-based model to detect sarcasm in Tamil code-mixed data. \cite{rosid2024sarcasm} consider the task of sarcasm detection in Indonesian-English code-mixed tweets and propose a hybrid model based on CNN with multi-head attention and bidirectional gated recurrent unit (BiGRU) with pragmatic auxiliary features. Similar studies \cite{anitha2024code,chauhan2024transformer} focus on both classical and transformer based approaches for code-mixed sarcasm detection in the Indic context. 


With the advent of LLMs, several studies have focused on detecting sarcasm through LLM based approaches as well. \cite{yao2025sarcasm} develop a step-by-step reasoning framework called \textit{SarcasmCue} that elicits LLMs to detect sarcasm through sequential and non-sequential prompting methods. \cite{chanda2024leveraging}, propose an approach to Tamil-English and Malayalam-English code-mixed sarcasm detection and perform a comparative analysis of XLM-RoBERTa and ChatGPT, for the task.

Our work is motivated by these previous studies, and aims to compare the performance of multiple LLMs (Phi 4, Llama-3.1, Gemma, and Mistral) in sarcasm detection in zero-shot and few-shot set ups with that of a classical transformer based model (DistilBERT), fine-tuned on domain specific text data. The current study is closest to that by \cite{chanda2024leveraging}. However, we perform a comparative analysis of multiple LLMs  with our DistilBERT model. Additionally, the previous study performs sarcasm detection in transliterated text, unlike code-mixed data. Another difference is that aligning with previous studies \cite{sravani2023enhancing,pratapa2018language,rizvi2021gcm,shravani2023unsupervised}, we propose a unique, LLM generated code-mixed Hinglish dataset for the task, which is used to fine-tune our transformer based model trained in English. Our approach thus intends to address the problem of sarcasm detection with minimal use of code-mixed data, ensuring its robustness in low-resource and data scarce set ups.

\section{Methodology}

\begin{figure*} [!htbp]
\small
	\centering
	\includegraphics[width=12.5cm]{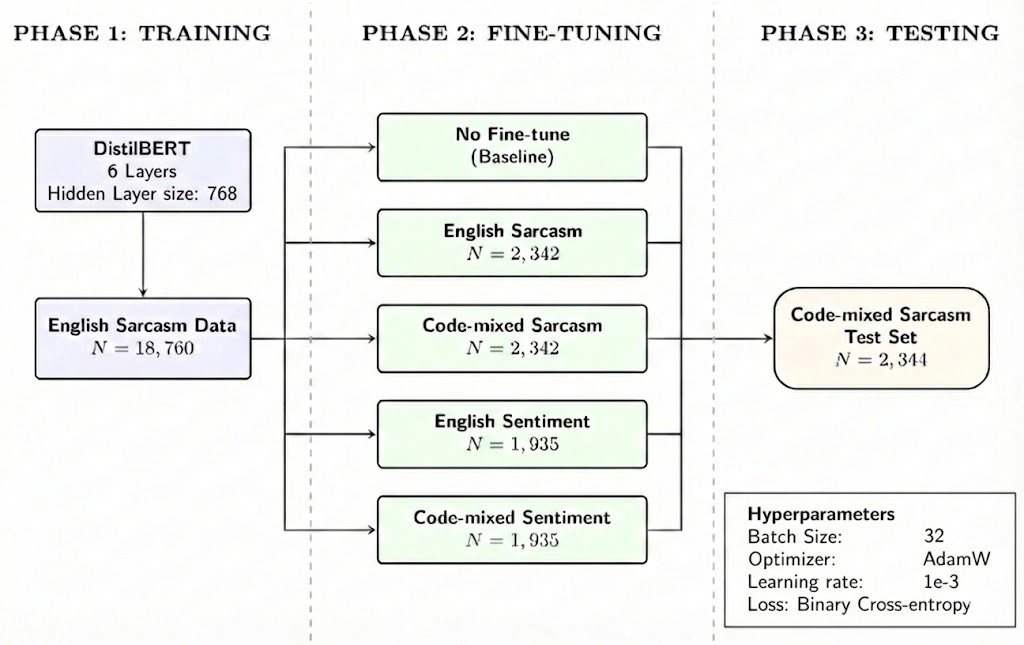}
	\caption{DistilBERT training and fine-tuning pipeline}
	\label{fig:block_diagram}
\end{figure*}

\subsection{Data}
Availability of code-mixed datasets on sarcasm, especially Hinglish, remains limited. To address this scarcity, we used two English datasets for this study: a sarcasm dataset \cite{misra2022newsheadlinesdatasetsarcasm} and a sentiment dataset \cite{go2009twitter}, which were translated into their Hinglish code-mixed counterparts using Google AI Studio. This was primarily done to enable a supervised learning set up; since the original English datasets are labeled, the same labels were carried forward to their code-mixed version.

We employed the \textbf{News Headlines Dataset for Sarcasm Detection}, which comprises over 18,000 instances of news headlines collected from two prominent online news sources: \textit{The Onion} and \textit{The HuffPost}. This dataset was selected due to its inherent suitability for sentence-level sarcasm classification, as the headlines are concise, self-contained, and context-independent. The dataset is annotated with two categorical labels: (i)\textit{sarcastic} and (ii)\textit{ non-sarcastic}. To explore the feasibility of leveraging sentiment analysis as an auxiliary task to enhance sarcasm detection via transfer learning, we also used the \textbf{Stanford Sentiment140} dataset as the sentiment dataset. This dataset contains 1.6 million tweets, out of which 1935 tweets were used for our task. This was primarily done to ensure minimal dependency on the availability of code-mixed fine-tuning data. Each data point in this dataset comprises three sentiment categories, namely \textit{positive}, \textit{negative}, and \textit{neutral}.

\subsection{Synthetic Code-Mixed Data Generation}\label{method:translation}
As stated earlier, we translated the English (sarcasm and sentiment) data into their code-mixed counterparts, to generate synthetic code-mixed data for the task. For the translation process, we utilized the Gemini 2.5 Pro \cite{comanici2025gemini} model available in Google AI Studio. The translation followed a trial-and-error prompting strategy aimed at generating high-quality code-mixed outputs. Particular emphasis was placed on preserving the semantic intent of the original sentences while representing Hindi lexical items in the Devanagari script within an otherwise English syntactic structure. After multiple iterations, a final prompt was selected that consistently produced linguistically coherent and semantically faithful Hinglish translations. The finalized prompt used for dataset generation is presented below  in Figure \ref{fig:prompt_translation}.

After the translation was performed, we carried out a manual check to see the translation quality. Two independent annotators (proficient in both English and Hindi) studied randomly selected 350 translations from both sarcasm and sentiment datasets, and marked each translation as \textit{satisfactory} and \textit{un-satisfactory}. A comprehensive analysis of these translation failures is provided in Appendix \ref{appendix:example_data}, which details specific instances where the model produced suboptimal results. This exercise led to an initial inter-annotator agreement of 85\%. Disagreements were finally resolved through discussion. We finally checked the number of dissatisfactory translations in this sample, which turned out to be 15\%. These results indicate that most translations were reasonably reliable, and thus, could be used for the fine-tuning and testing purposes. 

A small subset of sentences could not be processed due to model restrictions related to sensitive or contentious topics such as politics, body shaming, and warfare. These instances were manually translated to ensure dataset completeness and consistency. Table \ref{tab:example_translations} shows some example translations.

\begin{table}[!htbp]
\centering
\caption{Example English-to-Hinglish translations}
\label{tab:example_translations}
\renewcommand{\arraystretch}{1.2}
\begin{tabular}{p{0.45\columnwidth} p{0.45\columnwidth}}
\toprule
\textbf{Original Sentence} & \textbf{Translated Sentence} \\
\midrule
Bernie Sanders wins Maine democratic caucus & Bernie Sanders {\dn n\?} Maine democratic caucus {\dn Ejt\?}
\\
\midrule
Meet the white house's newest star: a whiteboard &  White house {\dn k\?} newest star {\dn s\?} {\dn Emlo}: {\dn ek} whiteboard\\
\midrule
Four consequences of a \$15 minimum wage & \$15 minimum wage {\dn k\?} {\dn cAr} consequences \\
\midrule
15 things to bring on your summer adventures & {\dn apn\?} summer adventures {\dn pr} {\dn l\?} {\dn jAn\?} {\dn k\?} {\dn Ela\?} 15 things\\
\bottomrule
\end{tabular}
\label{tab:translation}
\end{table}


\hspace{10pt}


\begin{figure}[t]
\centering
\begin{tcolorbox}[
    colback=white,
    colframe=black,
    boxrule=1pt,
    top=1.5mm,
    bottom=1.5mm,
    left=1mm,
    right=1mm,
    before skip=6pt,
    after skip=6pt
]
{\color{black}\textit{
``Translate the following sentences into Hindi-English code-mixed sentences. Use Hindi/Devanagari script for words written using Devanagari font, if any, in the original text. Example: ‘Weekend plans got cancelled.’ 
→ ‘Weekend plans cancel {\dn ho ge}' ''
}}
\end{tcolorbox}
\caption{Prompt used in Gemini 2.5 Pro for synthetic code-mixed data generation}
\label{fig:prompt_translation}
\end{figure}
\subsection{Data Preparation}
Both the original English datasets and their translated Hinglish code-mixed counterparts were partitioned into three subsets corresponding to training, fine-tuning, and testing. To mitigate potential class imbalance and reduce bias during model training, class distribution was carefully controlled across all subsets. Random undersampling was applied to ensure balanced representation of both classes within each data split.
The proportional distribution of instances across the training, fine-tuning, and testing sets is detailed in table \ref{tab: Sarc_data_dist}. The size of the fine-tuning dataset has deliberately been kept small, considering the low availability of code-mixed data on sarcasm in real world settings.
\begin{table}[!htbp]
\centering
\caption{Sarcasm dataset distribution}
\begin{tabular}{lcc}
\toprule
Dataset Split & Sarcastic & Non-Sarcastic \\
\midrule
Training & 9380 & 9380 \\
Fine-Tune (English) & 1171 & 1171 \\
Fine-Tune (Hinglish) & 1171 & 1171 \\
Testing & 1172 & 1172 \\
\bottomrule
\end{tabular}
\label{tab: Sarc_data_dist}
\end{table}

The size and class-wise distribution of the English sarcasm dataset and its code-mixed Hinglish (fine-tuning) counterparts were kept identical to ensure experimental consistency and enable a fair comparative analysis. The label-wise distribution of instances after pre-processing, for the fine-tuning sentiment data, is presented in table \ref{tab: sent_data_dist}.

\begin{table}[!htbp]
\centering
\caption{Sentiment dataset distribution}
\begin{tabular}{lccc}
\toprule
Split & Positive & Negative & Neutral \\
\midrule
Fine-tuning & 644 & 646 & 645 \\
\bottomrule
\end{tabular}
\label{tab: sent_data_dist}
\end{table}

\subsection{Training the Classical Model}\label{method:training}
We used \textit{DistilBERT-base-uncased} \cite{sanh2020distilbertdistilledversionbert}, which is a compact, distilled variant of the transformer-based language model \textit{Bidirectional Encoder Representations from Transformers (BERT)}, designed to perform with minimal computational overhead. Compared to its predecessor, the compressed DistilBERT model retains the majority of BERT’s semantic and contextual representation capabilities and comprises six transformer encoder layers equipped with multi-head self-attention mechanisms and position-wise feedforward networks.

Text preprocessing and tokenization were carried out using the corresponding DistilBERT tokenizer, which converts input sequences into token embeddings while maintaining case-insensitive consistency. Model training was conducted with a batch size of 32. The \textbf{AdamW} optimizer was used for parameter optimization, with a learning rate set to 1×$10^{-3}$. Training was performed for 100 epochs, using the binary cross-entropy loss. This allowed for loss convergence in all experimental set ups.

\subsection{Fine-tuning the Classical Model}
For fine-tuning purposes, sequential transfer learning was employed. Two fine-tuning paradigms were explored: one using minimal sarcasm-labeled data, and the other using minimal sentiment-labeled data. This dual approach enabled a comparative analysis of fine-tuning effectiveness and facilitated an empirical assessment of which auxiliary task contributes more effectively to sarcasm detection in code-mixed settings.

Fine-tuning was also carried out independently for the English and code-mixed Hinglish datasets across both sarcasm and sentiment classification tasks, enabling a controlled assessment of model performance across languages and affective dimensions.

Fine-tuning the model using sentiment data posed additional challenges due to a mismatch between the number of output labels in the sentiment classification task (three labels) and the target sarcasm detection task (two labels). To address this discrepancy, sequential transfer learning was adopted, wherein the learned weights from the pretrained model were retained and used as the initialization for subsequent fine-tuning. The label mismatch was resolved by modifying only the final classification layer, while preserving the learned representations in the lower layers of the network. The training and fine-tuning pipeline for the classical model is shown in figure \ref{fig:block_diagram}.

\subsection{LLM based Classification}
To evaluate and compare the effectiveness of LLMs in detecting sarcasm within code-mixed text, we conducted experiments using four state-of-the-art, pretrained open-source models: \textbf{LLaMA 3.1} \cite{grattafiori2024llama3herdmodels}, \textbf{Mistral} \cite{jiang2023mistral7b}, \textbf{Phi 4} \cite{abdin2024phi4technicalreport}, and \textbf{Gemma 3} \cite{gemmateam2025gemma3technicalreport}. These models were selected to represent diverse architectural designs and training paradigms within the contemporary LLM landscape.
All models were executed using \textbf{Ollama} \cite{ollama2023}, a locally deployed inference framework that enables efficient execution of open-source LLMs on local hardware. This setup ensured consistency in the evaluation environment while eliminating external API dependencies.

The prompting strategy adopted for this experiment was developed through a trial-and-error process, with the objective of eliciting accurate and consistent sarcasm classification responses from each model. The final prompt used for LLM-based evaluation is presented below in figure \ref{fig:prompt_llm_classify}.
\begin{figure}[t]
\centering
\begin{tcolorbox}[
    colback=white,
    colframe=black,
    boxrule=0.8pt,
    top=1mm,
    bottom=1mm,
    left=1mm,
    right=1mm,
    before skip=6pt,
    after skip=6pt
]
{\color{black}\textit{
``You are a sarcasm detection model. You have to detect sarcasm in Hinglish sentences. 
Sentence: "{sentence}"
Don't give any explanation and Respond ONLY with one label:
- Sarcastic
- Non-Sarcastic''
}}
\end{tcolorbox}
\caption{Prompt used for LLM-based sarcasm classification}
\label{fig:prompt_llm_classify}
\end{figure}
Here, the term sentence refers to an individual textual instance from the dataset. Each sentence was processed independently and sequentially, with a separate API call for every input instance. 

\section{Results}
The sarcasm classification experiments were conducted using four LLMs, alongside DistilBERT as a strong transformer-based model. For all DistilBERT experiments, the model was trained exclusively on English sarcasm data. This design choice minimizes reliance on code-mixed sarcasm datasets, which are typically limited in availability. To ensure a fair and controlled comparison, the same class-balanced code-mixed testing dataset was employed for both the LLM-based evaluations and the DistilBERT experiments.

\subsection{LLM Performance}
\begin{table*}[t]
\centering
\caption{LLM performance for sarcasm detection under zero-shot and few-shot prompting.}
\setlength{\tabcolsep}{6pt}
\renewcommand{\arraystretch}{1.15}
\begin{tabular}{lcccccccc}
\toprule
\multirow{2}{*}{Models} & \multicolumn{4}{c}{Zero-Shot} & \multicolumn{4}{c}{Few-Shot} \\
\cmidrule(lr){2-5} \cmidrule(lr){6-9}
 & \multicolumn{2}{c}{English} & \multicolumn{2}{c}{Code-mixed} & \multicolumn{2}{c}{English} & \multicolumn{2}{c}{Code-mixed} \\
\cmidrule(lr){2-3} \cmidrule(lr){4-5} \cmidrule(lr){6-7} \cmidrule(lr){8-9}
 & Accuracy & F1-Score & Accuracy & F1-Score & Accuracy & F1-Score & Accuracy & F1-Score \\
\midrule
Llama 3.1 & 0.6681 & 0.66 & 0.4953 & 0.34 & 0.6907 & 0.68 & 0.6229 & 0.60 \\
Mistral   & 0.6079 & 0.55 & 0.5341 & 0.42 & 0.5328 & 0.41 & 0.5030 & 0.34 \\
Gemma 3   & 0.6553 & 0.60 & 0.5806 & 0.53 & 0.5034 & 0.34 & 0.5614 & 0.47 \\
Phi 4     & 0.6945 & 0.69 & 0.6036 & 0.55 & 0.6540 & 0.62 & 0.6028 & 0.56 \\
\bottomrule
\end{tabular}
\label{tab:LLM_pred}
\end{table*}
Sarcasm detection experiments were conducted using the LLMs in zero-shot and few-shot set ups. The quantitative evaluation results are summarized in table \ref{tab:LLM_pred}. In the zero-shot setting, Phi 4 demonstrated the highest performance for code-mixed test data, achieving an accuracy of 60.36\% (the confusion matrix is shown in Table \ref{appendix:phi4_matrix}), whereas the open-source LLM Llama 3.1 exhibited the lowest accuracy, at 49.53\%. This performance disparity highlights notable variations in sarcasm detection capability across architectures.

In the few-shot setting, however, Llama 3.1 demonstrated the highest accuracy (confusion matrix in the appendix), surpassing Phi 4 by two percentage points. This also shows that the superior/inferior performance of LLMs in the zero-shot setting may not always propagate to the few-shot setting, for the given data. The best and the worst performing models remain the same for English test data. Notably, the performance of all LLMs on English is significantly better than that on code-mixed data\footnote{For each model, we performed a paired bootstrap significance test (2344 resamples) comparing accuracy on English and Hinglish test sets. Across all models, English consistently outperformed Hinglish, with the 95\% confidence interval of the accuracy difference excluding zero.}; an artifact of predominant English pre-training of the LLMs.  


\subsection{DistilBERT Performance}
The DistilBERT-based experiments were conducted following a set of distinct fine-tuning strategies to analyze the impact of different fine-tuning schemes as listed below. The same Hinglish code-mixed test data was used in each set up:
\begin{itemize}
    \item \textbf{No fine-tuning:} The model was trained on English training data without any fine-tuning.
    \item \textbf{With sarcasm fine-tuning:} The model was trained on English data, and fine-tuned on (A) English or (B) Code-mixed sarcasm data.
    \item \textbf{With sentiment fine-tuning:} The model was trained on English data, and fine-tuned on (A) English or (B) Code-mixed sentiment data.
\end{itemize}
We performed 5-fold cross-validation on the English training set for hyperparameter selection (average training accuracy of 92.75\% was obtained, indicating effective learning and consistent convergence across folds). The final model was then retrained on the full training data using the selected configuration and evaluated on the held-out test set (hyperparameters in section \ref{method:training}). Table \ref{tab:distilbert_results} reports the corresponding test accuracy and F1 scores obtained under each approach.


Among these strategies, the approach involving pre-training on English sarcasm data followed by fine-tuning on code-mixed sarcasm data emerged as the most effective, achieving an accuracy of approximately 84\% (The confusion matrix of results for this experiment is in Table \ref{appendinx:ft_cm_sarc}).  In contrast, the \textit{English Sentiment fine-Tuning} strategy, in which the model was trained on English sarcasm, fine-tuned on English sentiment data, and tested on code-mixed sarcasm data, resulted in the poorest performance, with an accuracy of approximately 54\%. 

\begin{table}[!htbp]
    \centering
    \caption{DistilBERT result summary: Code-mixed fine-tuning leads to best performance}
    \begin{tabular}{p{2cm}|p{2cm}|p{1.6cm}|c}
        \textbf{Fine-tuning data}&\textbf{Testing data}&\textbf{Accuracy}&\textbf{F1 Score}  \\
        \midrule
        No fine-tune & Code-mixed [Sarcasm] & 0.56868 & 0.47\\
        \midrule
        English [Sarcasm] & Code-mixed [Sarcasm] & 0.64462 & 0.61 \\
        \midrule
        Code-mixed [Sarcasm] & Code-mixed [Sarcasm] & \textbf{0.83873} & \textbf{0.84}\\
        \midrule
        English [Sentiment] & Code-mixed [Sarcasm] & 0.53882 & 0.47\\
        \midrule
        Code-mixed [Sentiment] & Code-mixed [Sarcasm] & 0.58873 & 0.59\\
        \midrule
    \end{tabular}
    \label{tab:distilbert_results}
\end{table}

In contrast to the above par performance in code-mixed fine-tuning, English fine-tuning resulted in a considerably lower test accuracy of 64.46\%. These results further underscore the importance of language-specific adaptation, particularly for code-mixed sarcasm detection tasks.

\begin{figure}[]
    \centering
    \includegraphics[width=9cm]{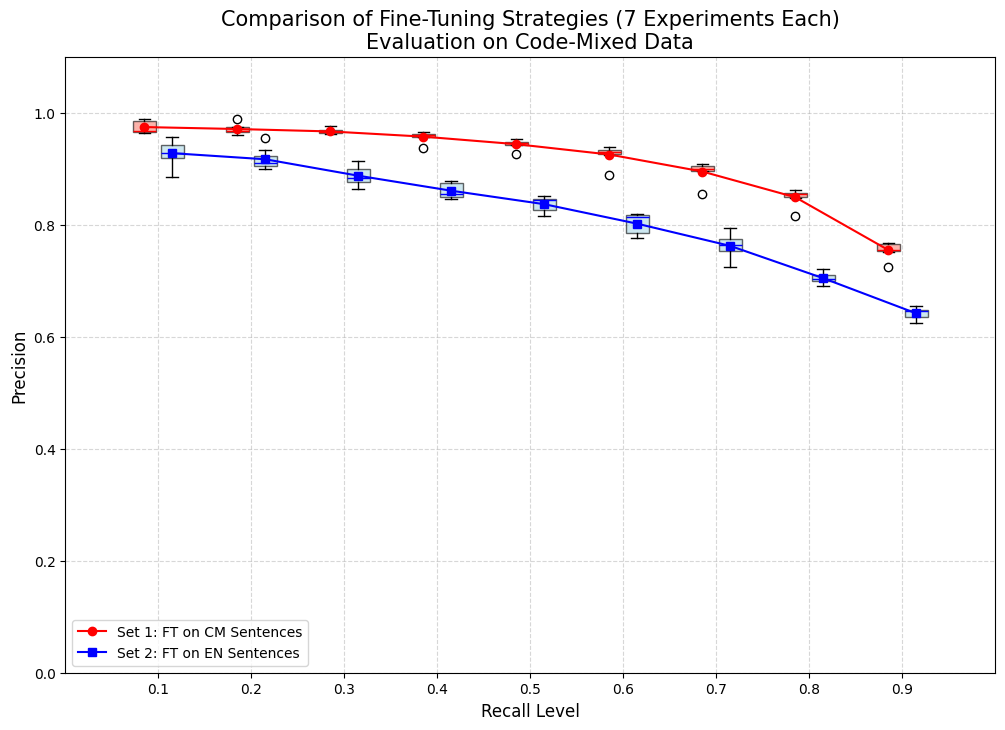}
    \caption{AUPRC Curve after Fine-tuning with Code-mixed Sarcasm (red) and English Sarcasm (blue) data}
    \label{fig:auprc}
\end{figure}

Figure \ref{fig:auprc} presents the Precision–Recall (PR) curves evaluating model performance on code-mixed data under two fine-tuning strategies: code-mixed (CM) and English (EN) sentences. When fine-tuned on code-mixed sarcasm, the model achieves a substantially higher average Area Under the Precision–Recall Curve (AUPRC) of 0.9012 across seven experiments. This reflects a robust and consistent trade-off between precision and recall across varying decision thresholds. Notably, the CM fine-tuned model sustains high precision levels (approximately 0.85) even up to a recall of 0.80, indicating the effective identification of sarcastic instances while maintaining a low false-positive rate.

In contrast, fine-tuning exclusively on English sentences results in a comparatively lower average AUPRC of 0.8080. Under this setting, precision exhibits a steady, near-linear decline as recall increases, dropping to approximately 0.65 at a recall level of 0.90. This trend suggests diminished class separability on the target code-mixed evaluation data, highlighting a higher tendency to misclassify non-sarcastic instances as sarcastic when the model lacks prior exposure to code-mixed structures during fine-tuning.

When compared to sarcasm based fine-tuning, sentiment fine-tuning yielded significantly poorer results (53.88\% for English and 58.87\% for code-mixed). The confusion matrix for Code-mixed Fine-tuning is in Table \ref{appendix:matrix_cm_sent} and the matrix for English Fine-tuning is in Table \ref{appendix:matrix_en_sent}.However, in this case too, we observe an advantage of the code-mixed fine-tuning approach, over English fine-tuning.

It can thus be observed that the DistilBERT model significantly dominates the LLMs in sarcasm detection on our dataset (for each model, we performed a paired bootstrap significance test comparing accuracy of both DistilBERT and the best Few-shot fine-tuning model on the Hinglish test set. Across all models, English consistently outperformed Hinglish, with the 95\% confidence interval of the accuracy difference excluding zero). In particular, fine-tuning with code-mixed sarcasm data yields the highest performance, indicating the model’s enhanced ability to capture linguistic and contextual nuances inherent in code-mixed expressions. Although fine-tuning with English sarcasm data results in comparatively lower performance than the code-mixed setting, it still achieves accuracy levels that surpass those of the evaluated LLMs, underscoring the effectiveness of task-specific fine-tuning in comparison to zero-shot or prompt-based LLM inference.

\subsection{Training Data Size Versus Model Performance}
\begin{figure}[!htbp]
    \centering
    \includegraphics[width=1\linewidth]{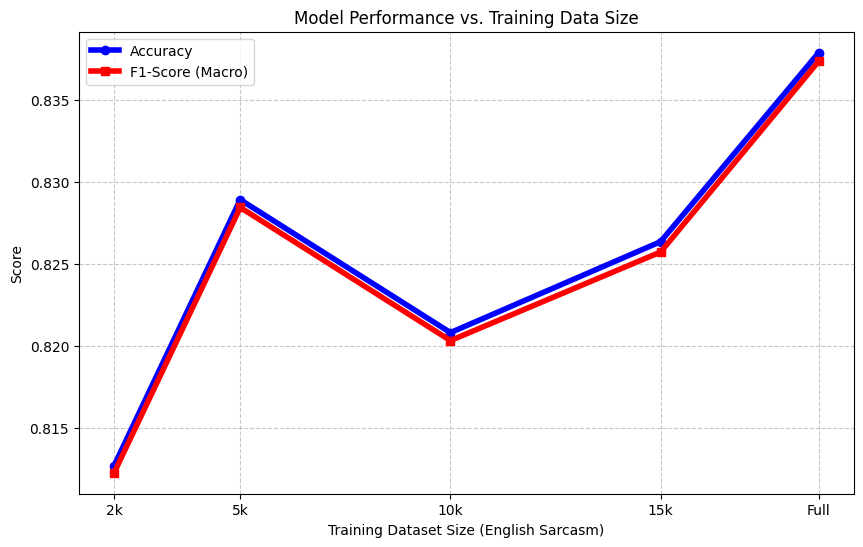}
    \caption{Model performance with respect to different Training Size before Fine-tuning on Code-Mixed Sarcasm data}
    \label{fig:TrainSize_FT_CM}
\end{figure}

\begin{figure}[!htbp]
    \centering
    \includegraphics[width=1\linewidth]{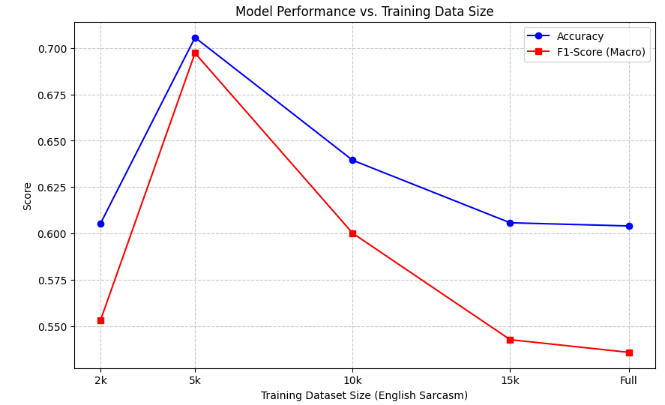}
    \caption{Model performance with respect to different Training Size before Fine-tuning on English Sarcasm data}
    \label{fig:Trainsize_FT_EN}
\end{figure}

To further analyze the impact of the scale of training data, the two best-performing approaches identified in the earlier experiments (code-mixed and English sarcasm fine-tuning) were selected for evaluation under varying training set sizes (figures \ref{fig:TrainSize_FT_CM} and \ref{fig:Trainsize_FT_EN}). The model was trained on progressively larger subsets of data and subsequently fine-tuned on the same amount of code-mixed sarcasm and English sarcasm datasets (as in the previous section), respectively.

For the code-mixed sarcasm fine-tuning strategy (figure \ref{fig:TrainSize_FT_CM}), the model exhibits a clear performance improvement as the training dataset size increases. Although a temporary decline is observed around the 10K sample point, the overall trend remains positive, with the model achieving its highest Accuracy and F1-score when trained on the full dataset, indicating the impact of larger and more diverse training data on detecting code-mixed sarcasm.

In contrast, when the model is fine-tuned on English sarcasm data (figure \ref{fig:Trainsize_FT_EN}), both accuracy and F1-score show a consistent decline beyond a certain threshold, notably after the 5K sample point. Moreover, the reduction in Macro F1-score (from 0.69 to 0.6) is more pronounced than the corresponding drop in accuracy f(0.71 to 0.63), indicating a deterioration in class-wise detection performance. This trend reflects reduced performance on minority classes, highlighting the limitations of English-only fine-tuning for code-mixed sarcasm detection.

\section{Discussion}
Our experiments reveal the superiority of domain specific fine-tuned model over SOTA LLMs, in the task of sarcasm detection. We see that the DistilBERT model trained on English sarcasm data (without fine-tuning) demonstrates strong performance in detecting sarcasm within English sentences. However, a noticeable degradation in performance occurs when the model is evaluated on code-mixed inputs (accuracy: 56.87\%). This performance decline can be attributed primarily to the altered syntactic and lexical structure inherent to code-mixed language, which differs substantially from monolingual English text and poses challenges for models trained exclusively on standard English corpora.

Fine-tuning the base DistilBERT model using English-language sarcasm data results in a significant improvement in accuracy (64.46\%). This performance already significantly surpasses the best performing LLM across both zero-/few-shot set ups (Llama 3.1 few-shot: 62.29\%). This finding highlights the effectiveness of domain-specific fine-tuning of classical transformer models, demonstrating that task-adapted models can even outperform substantially larger LLMs in sarcasm detection, even when the training and fine-tuning phases are devoid of code-mixed data.  

Fine-tuning the model with Hinglish sarcasm data led to the best performance in sarcasm detection. The model achieves an accuracy of 83.87\% (outperforming the best LLM by nearly 20 percentage points), indicating a significant benefit from language- and domain-specific adaptation. Additionally, the classical model achieves this performance with minimal use of LLM generated code-mixed fine-tuning data. This is especially significant, since code-mixed data around highly specific tasks like sarcasm detection have limited availability. The results also pave pathways towards exploration of LLM based synthetic data generation, in data scarce tasks.

The motivation for incorporating sentiment data through transfer learning stemmed from the close conceptual relationship between sentiment and sarcasm, with sarcasm often viewed as a nuanced or implicit form of sentiment expression. Despite this theoretical alignment, fine-tuning with sentiment data did not yield substantial performance gains. In both scenarios, i.e., fine-tuning with English sentiment data and with code-mixed Hinglish sentiment data, the model exhibited a below-par performance, comparable to the base model without fine-tuning (53.88\% for the English sentiment fine-tuned model, and 58.87\% for the code-mixed sentiment fine-tuned model). This indicates that sentiment-level supervision alone may be insufficient for effective sarcasm detection. However, it must be noted that the benefit of fine-tuning the model with code-mixed sentiment data can still be observed, with it outperforming its English fine-tuned counterpart by 5 percentage points, and achieving a performance close to the best performing LLM in sarcasm detection.

\textbf{Limitations and Future Work:} We acknowledge that our findings hold specifically for the datasets used in the study. As part of future work, we plan to extend this study to multiple languages in order to better capture linguistic diversity and a wider range of code-mixing patterns. Inclusion of more regional Indian languages in the study will make our findings robust and more generalizable. Another important direction to explore would be inclusion of publicly available web-based data (social media data and blogs) to measure the robustness of the findings in real-world scenarios.

Our mixed-method validation involves manual qualitative analysis, which is inherently subjective and may yield different outcomes under more rigorous annotation settings. We mitigate this through consistent annotation guidelines and plan to incorporate multiple annotators in future work to ensure reliability and robustness of our datasets and findings.

Our results around transfer learning through sentiment classification being used as an auxiliary task for sarcasm detection, also requires further inspection. While the sentiment data currently used to fine-tune the model is small, we plan to experiment further with this set up, by expanding the code-mixed sentiment dataset and observing its effect on the model's performance.

We also perform an analysis of the misclassifications (see section \ref{appendix:analysis_misclassif} in the appendix). We see that the lack of context stemming from the complexity of translating a US based dataset into its equivalent Indic code-mixed counterpart leads to a significant number of dissatisfactory translations, and misclassifications by the best performing model. We plan to extend this work to sarcasm datasets attuned to the Indian context, for a comparative analysis. 

Finally, a major next step of comparing the performance of the classical model (DistilBERT) with fine-tuned LLMs will further our work and provide deeper insights into the trade-offs between model capacity, task-specific adaptation, and performance.

\section{Conclusion}
In this paper, we perform a comparative analysis of the performance of a domain fine-tuned classical transformer based model and four SOTA LLMs, for the task of sarcasm detection in code-mixed Hinglish text. We find that the smaller, domain fine-tuned, classical model significantly outperforms the LLMs in both zero and few-shot set ups, especially when fine-tuned on minimal code-mixed data generated by an LLM. The classical model, even when fine-tuned on a different yet related auxiliary task of sentiment classification, exhibits commendable performance comparable to the LLM with the best performance in sarcasm detection. These findings highlight the importance of domain adaptation of smaller models, especially in scenarios with data scarcity. Our work serves as a formative step toward advancing sarcasm detection in code-mixed settings and exploring the use of LLM-based data augmentation for data-scarce NLP tasks.

\bibliography{custom}

\appendix
\section{Appendix}
\subsection{Examples of Code-Mixed Sentences}\label{appendix:example_data}
To provide a comprehensive view of the dataset and illustrate the linguistic nuances of code-mixing, this appendix presents curated samples of both sarcastic and non-sarcastic utterances. Each English source sentence is paired with its corresponding code-mixed (Hindi-English) translation. This comparative mapping highlights the distinct syntactic and semantic adaptations required when processing different types of discourse in a multilingual context.

Table \ref{fig:sarc_cm_ex} details ten instances of sarcastic sentences. Translating cross-lingual irony presents a unique challenge, as the system must preserve the underlying ironic intent and tone while navigating linguistic boundaries. The provided examples demonstrate how the model integrates English colloquialisms and specific entities (e.g., ``down in dumps," ``TSA agents," ``urine tank", etc.) directly into the Hindi syntactic structure. This selective retention of source vocabulary acts as a mechanism to maintain the original comedic or satirical intent within the code-mixed output. However, at times, this same feature also results in loss of context and unrealistic code-mixed equivalents. We discuss these details in the next section.
\begin{table}[!htbp]
 \caption{Sarcastic code-mixed sentences misclassified by the best performing model}
    \centering
    \includegraphics[width=\linewidth]{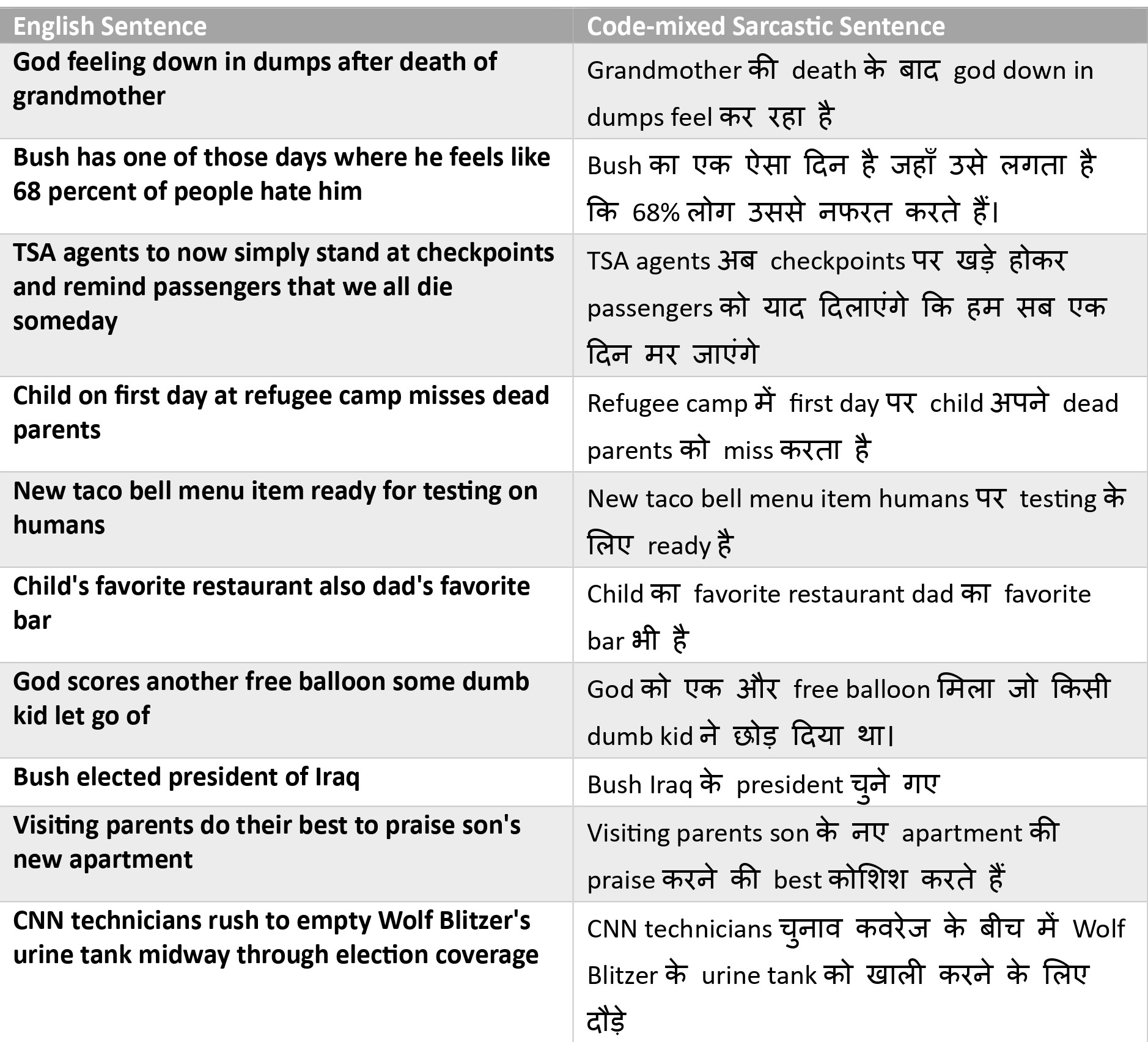}
    \label{fig:sarc_cm_ex}
\end{table}

Table \ref{fig:nonsarc_cm_ex} provides a sample consisting of ten non-sarcastic sentences. In these instances, the translation process is more straightforward, focusing strictly on factual preservation and semantic fidelity without the added complexity of figurative language or sarcastic intent. 
\begin{table}[!htbp]
\caption{Non-Sarcastic code-mixed sentences}
    \centering
    \includegraphics[width=\linewidth]{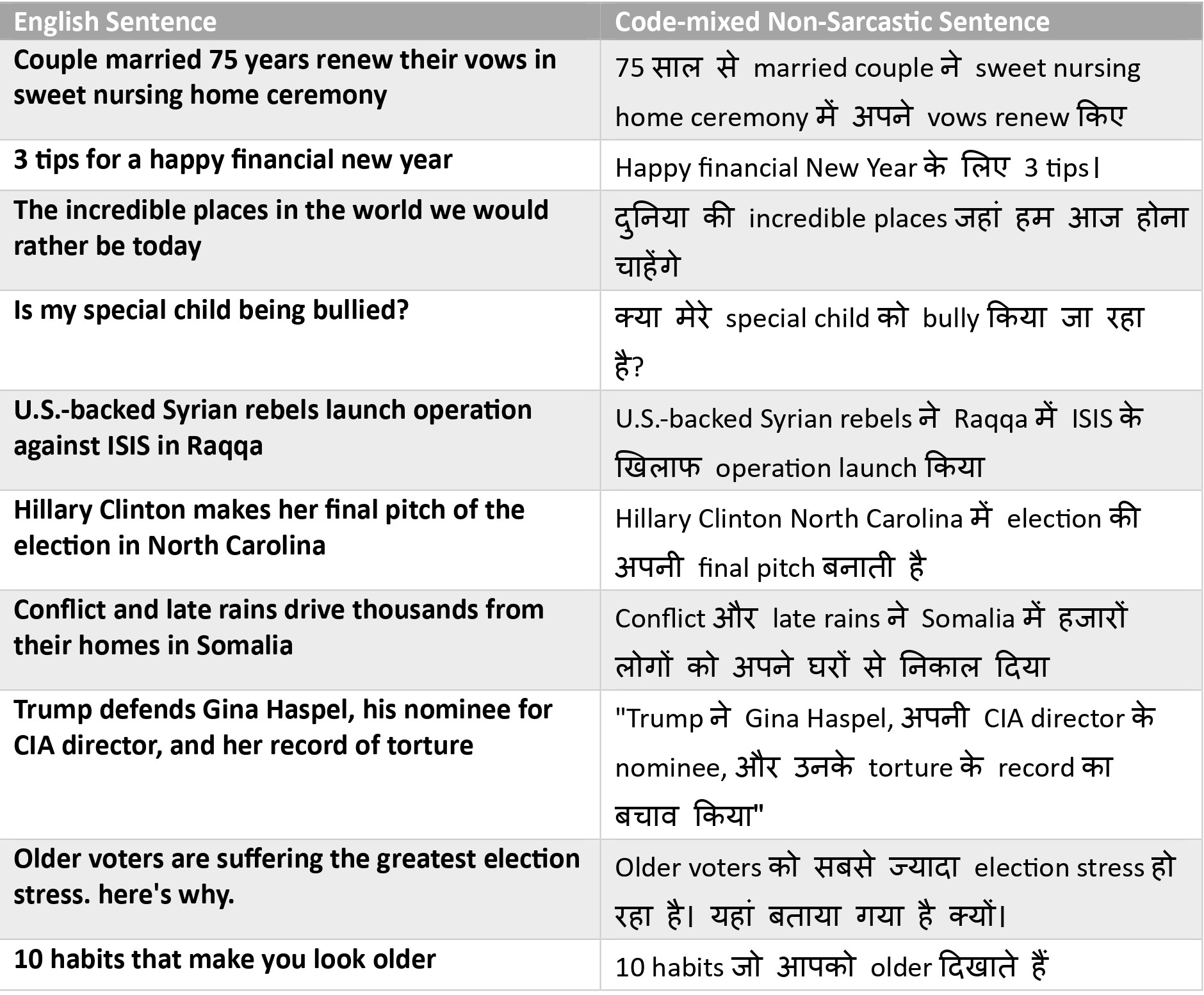}
    \label{fig:nonsarc_cm_ex}
\end{table}
\subsection{Quality of Translation by LLMs}\label{appendix:quality_translate}
As discussed in section \ref{method:translation}, the overall translation quality produced by the Gemini 2.5 model demonstrates a high degree of accuracy. However, in some cases the syntactic structure and semantics of the source text are distorted during the translation process. In other words, we observe a significant number of cases where the translation leads to code-mixed sentences that do not align well with natural use of the two languages. This outcome is expected, as the original dataset is predominantly grounded in a US-centric context. Therefore, the LLM encounters difficulty in identifying semantically equivalent Indic expressions for several colloquial US terms and slang.

As illustrated in table \ref{fig:example_translations}, these suboptimal outcomes appear due to overtly literal translations that obscure the underlying nuances of the original text. The translation model additionally exhibits a tendency to retain complex, low-frequency source words/phrases untranslated (e.g., words like \textit{entire, white castle, pathetic etc}. This disrupts the structural coherence of the output with the original sentence, leading to semantically ambiguous code-mixed (Hinglish) data points.
\begin{table}[t]
\caption{Examples of translated (code-mixed) sentences, with manually annotated quality of translation}
\includegraphics[width=1.0\linewidth]{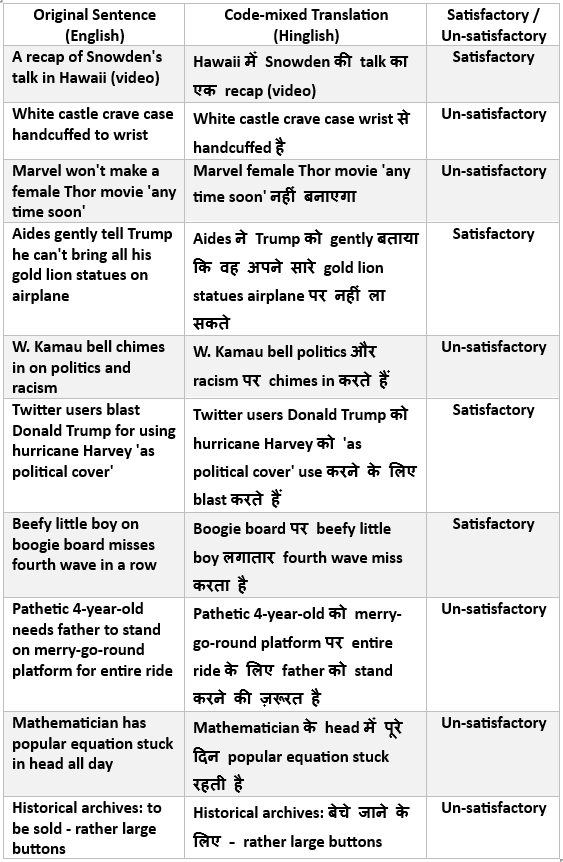}
    \label{fig:example_translations}
\end{table}
On the other hand, successful translations demonstrate a more balanced and context-aware approach. In these instances, the model selectively retains only high-frequency words/phrases commonly utilized in everyday colloquial discourse, thereby preserving the fluency and natural cadence of the translated text. 

\subsection{Confusion Matrices}
\subsubsection{LLM Performance}\label{appendix:phi4_matrix}
Table \ref{tab:conf_matrix_phi4} shows the performance of the Phi 4 model (best performing zero-shot model) in identifying Code-mixed Sarcasm sentences. The matrix shows that even though the recall is high, it suffers from low precision score, indicating the challenge it faces in sarcasm detection.
\begin{table}[!htbp]
\caption{Confusion matrix for Phi 4 (zero-shot)}
\label{tab:conf_matrix_phi4}
\begin{subfigure}{0.45\textwidth}
        \centering\small
        \begin{tabular}{c|c|c}
         & Pred. Sarc & Pred. Non-Sarc \\ \hline
        Actual Sarc & 1103 & 69 \\
        Actual Non-Sarc & 860 & 312 \\
        \end{tabular}
    \end{subfigure}
\end{table}
Additionally, table \ref{tab:conf_matrix_llama3.1} demonstrates the performance of best performing Llama 3.1 in the few-shot setup. Here we see a significantly higher precision, with a poor recall.
\begin{table}[!htbp]
\caption{Confusion matrix for Llama 3.1 (few-shot)}
\label{tab:conf_matrix_llama3.1}
\begin{subfigure}{0.45\textwidth}
        \centering\small
        \begin{tabular}{c|c|c}
         & Pred. Sarc & Pred. Non-Sarc \\ \hline
        Actual Sarc & 474 & 698 \\
        Actual Non-Sarc & 186 & 986 \\
        \end{tabular}
    \end{subfigure}
\end{table}

\subsubsection{Sentiment Fine-tuned DistilBERT}
Table \ref{appendix:matrix_en_sent} describes the classical model's performance in sarcasm detection on code-mixed data, after fine-tuning on English Sentiment. The high recall indicates the model's ability to detect most sarcastic sentences correctly. However, this comes at the cost of precision. Table \ref{appendix:matrix_cm_sent} shows the performance of the model fine-tuned on code-mixed Sentiment data. The precision is slightly better in this case, with a sacrifice of recall.
\begin{table}[!htbp]
\caption{Confusion matrices for sentiment based transfer learning using DistilBERT}
\centering
\begin{subfigure}{0.45\textwidth}
        \centering\small
        \begin{tabular}{c|c|c}
         & Pred. Sarc & Pred. Non-Sarc \\ \hline
        Actual Sarc & 1054 & 118 \\
        Actual Non-Sarc & 963 & 209 \\
        \end{tabular}
        \caption{Sentiment English fine-tuning}
        \label{appendix:matrix_en_sent}
    \end{subfigure}
\hspace{10pt}
\begin{subfigure}{0.45\textwidth}
        \centering\small
        \begin{tabular}{c|c|c}
         & Pred. Sarc & Pred. Non-Sarc \\ \hline
        Actual Sarc & 775 & 397 \\
        Actual Non-Sarc & 567 & 605 \\
        \end{tabular}
        \caption{Sentiment Code-mixed fine-tuning}
        \label{appendix:matrix_cm_sent}
    \end{subfigure}
\end{table}

\subsubsection{Code-mixed Sarcasm Fine-tuned DistilBERT}\label{appendinx:ft_cm_sarc}
The confusion matrix in table \ref{tab:bestmodelcm} shows the performance of the classical model fine-tuned on code-mixed Sarcasm data. It is evident that an optimal trade-off between precision and recall results in this model being outperforming all other models.
\begin{table}[!htbp]
\caption{Confusion matrix for the best performing DistilBERT model}
    \begin{subfigure}{0.45\textwidth}
        \centering\small
        \begin{tabular}{c|c|c}
         & Pred. Sarc & Pred. Non-Sarc \\ \hline
        Actual Sarc & 909 & 263 \\
        Actual Non-Sarc & 115 & 1057 \\
        \end{tabular}
    \end{subfigure}
    \label{tab:bestmodelcm}
\end{table}

\subsection{Analysis of Misclassified sentences}\label{appendix:analysis_misclassif}
A detailed error analysis of the misclassified instances reveals distinct patterns across models. For the best performing DistilBERT model fine-tuned on code-mixed sarcasm data, the majority of misclassified sentences exhibit limited contextual cues, making the sarcastic intent ambiguous (e.g., sentences 2 and 3 in table \ref{fig:misclassif_example}). In many such cases, the utterances rely heavily on implicit context or shared background knowledge, which is not explicitly available in the text. Consequently, the model struggles to infer the intended irony due to insufficient semantic or pragmatic signals.

In contrast, the LLMs demonstrate a different error profile. Their misclassifications frequently involve sentences containing offensive or sensitive vocabulary, including slang expressions and references to violence, like in the sentences 5 and 6 of table \ref{fig:misclassif_example}. Additionally, sentences mentioning political entities are repeatedly misclassified, suggesting potential sensitivity or bias in handling politically charged content (e.g., sentence 1 in table \ref{fig:misclassif_example}). In several instances like in sentence 7 of table \ref{fig:misclassif_example}, literal translations from English to a code-mixed form results in semantically distorted sentences, leading to a loss of the underlying sarcastic intent. Such translation-induced issues adversely affect both LLMs and the classical model, causing failures in accurate sarcasm detection.

\begin{table}[!htbp]
\caption{Examples of Misclassified Sentences by the best performing model (DistilBERT fine-tuned on code-mixed)}
    \centering
    \includegraphics[width=0.8\linewidth]{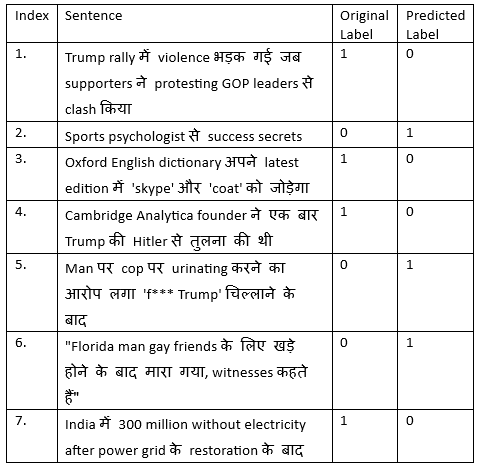}
    \label{fig:misclassif_example}
\end{table}

\end{document}